\documentclass[conference]{IEEEtran}
\IEEEoverridecommandlockouts
\usepackage{cite}
\usepackage{amsmath,amssymb,amsfonts}
\usepackage{graphicx}
\usepackage{textcomp}
\usepackage{xcolor}

\usepackage{algpseudocode}
\usepackage{url}
\usepackage{float}
\usepackage{booktabs}
\def\BibTeX{{\rm B\kern-.05em{\sc i\kern-.025em b}\kern-.08em
    T\kern-.1667em\lower.7ex\hbox{E}\kern-.125emX}}
\begin{document}

\title{ARMARecon: An ARMA Convolutional Filter based Graph Neural Network for Neurodegenerative Dementias Classification\\
 \thanks{IEEE International Symposium on Biomedical Imaging (ISBI) 2026}
}

\author{
\IEEEauthorblockN{VSS Tejaswi Abburi \IEEEauthorrefmark{1}, Ananya Singhal\IEEEauthorrefmark{2} ,Saurabh J. Shigwan\IEEEauthorrefmark{1} and Nitin Kumar\IEEEauthorrefmark{1}}
\IEEEauthorblockA{\IEEEauthorrefmark{1}\textit{Shiv Nadar Institution of Eminence, Delhi NCR, India}}
\IEEEauthorblockA{\IEEEauthorrefmark{2}
\textit{GE Healthcare Banglore}\\
\{va797, as146, saurabh.shigwan, nitin.kumar\}@snu.edu.in}
}

\maketitle

\begin{abstract}  
Early detection of neurodegenerative diseases such as Alzheimer’s Disease (AD) and Frontotemporal Dementia (FTD) is essential for reducing the risk of progression to severe disease stages.  As AD and FTD propagate along white-matter regions in a global, graph-dependent manner, graph-based neural networks are well suited to capture these patterns. Hence, we introduce ARMARecon, a unified graph learning framework that integrates Autoregressive Moving Average (ARMA) graph filtering with a reconstruction-driven objective to enhance feature representation and improve classification accuracy. ARMARecon effectively models both local and global connectivity by leveraging 20-bin Fractional Anisotropy (FA) histogram features extracted from white-matter regions, while mitigating over-smoothing. Overall, ARMARecon achieves superior performance compared to state-of-the-art methods on the multi-site dMRI datasets ADNI and NIFD.
Code is available at \url{https://github.com/tejaswi-abburi1083/ISBI2026-ARMARecon.git}.  
\end{abstract}  

\section{Introduction}  
Deep learning techniques have advanced disease diagnosis by automating the analysis of complex imaging data, especially in neurological disorders. This is crucial for conditions such as Frontotemporal Dementia (FTD) and Alzheimer’s Disease (AD), where early detection greatly improves treatment effectiveness.  

Early diffusion MRI studies depended on handcrafted features like Fractional Anisotropy (FA) and used traditional classifiers such as Support Vector Machines (SVM)~\cite{cortes1995support}, Random Forest~\cite{breiman2001random}, and XGBoost~\cite{chen2016xgboost}. While these methods achieved moderate accuracy~\cite{wen2021reproducible}, they could not capture the complex relationship across individual subjects in and across different cohorts\cite{chen2023tractgraphcnn}. To tackle this issue, graph-based methods—like Graph Convolutional Networks (GCNs)~\cite{kipf2017semi}, Graph Attention Networks (GATs)~\cite{veličković2018graphattentionnetworks}, Chebyshev Networks (ChebNets)~\cite{defferrard2016convolutional}, Autoencoder-constrained GCNs (AE-GCNs)~\cite{ma2021aegcn} were introduced. However, these models often have trouble capturing both local and global graph context, limiting their ability to generalize across different subjects and datasets.  

To address these challenges, we propose \textbf{ARMARecon}, a unified framework that mixes ARMA graph filtering~\cite{bianchi2021graph} with a reconstruction based regularizer to improve feature robustness and classification performance in diffusion MRI-based analysis of neurodegenerative diseases. Our method builds a complete subject-level graph where each node represents an individual’s FA-based feature vector, allowing nodes to share information through neighborhood aggregation. This helps capture population-level disease progression patterns that develop gradually rather than as distinct classes.  

The proposed model combines an ARMA convolution network~\cite{bianchi2021graph} with a reconstruction based regularizer~\cite{wang2017mgae} to reduce the over-smoothing problem often seen in GNNs. The ARMA filter combines autoregressive and moving-average components to model long-range dependencies and richer graph dynamics without adding too many parameters.  

The key contributions of this work are summarized as follows:  
\begin{itemize}  
    \item We introduce \textbf{ARMARecon}, a diffusion MRI-based graph learning framework that uses ARMA graph filtering with a reconstruction-driven learning objective for early detection of neurodegenerative diseases such as AD and FTD.  
    \item The proposed model effectively captures both local and global connectivity patterns while remaining strong against over-smoothing and label scarcity.  
    \item Comprehensive experiments on multi-site diffusion MRI datasets (ADNI and NIFD) show that ARMARecon outperforms traditional machine learning and existing deep learning architectures in accuracy and generalization.  
\end{itemize}  
\begin{figure*}[!htbp]
    \centering
    \includegraphics[width=\linewidth, trim={0 12 0 0}, clip]{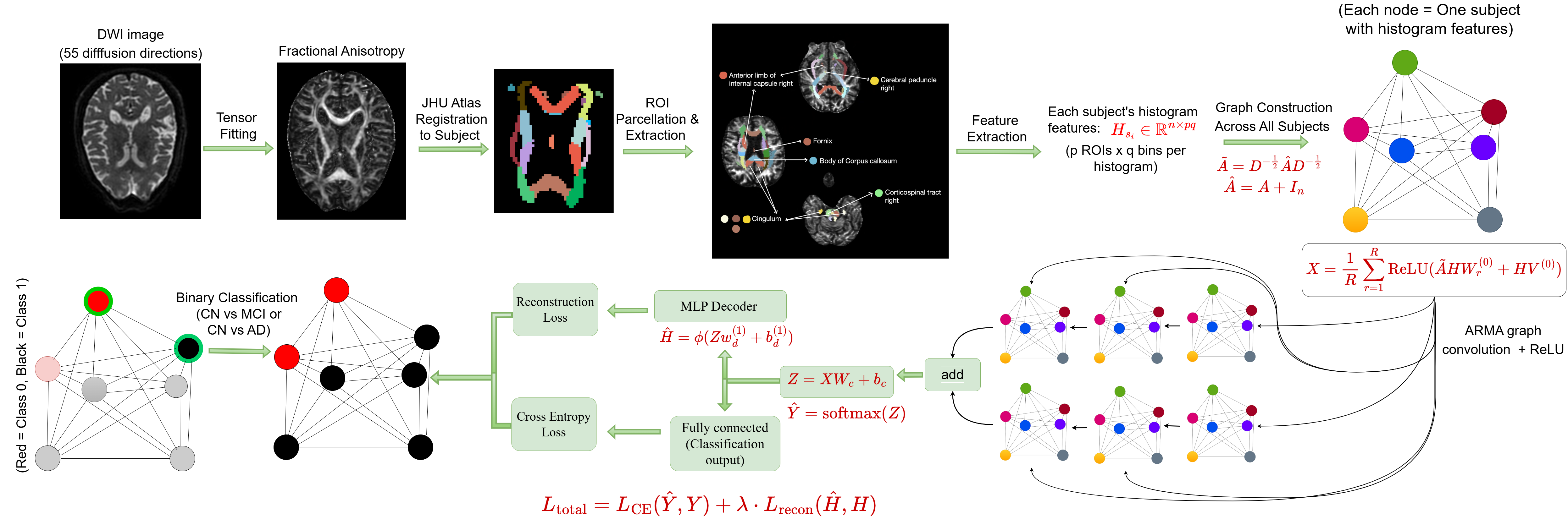}
    \caption{Proposed ARMARecon architecture integrating ARMA graph filtering with reconstruction regularization.}
    \label{fig:arma_architecture}
\end{figure*}


\section{Proposed Method}
\noindent
As shown in figure \ref{fig:arma_architecture}, in the preprocessing step we followed the standard procedure of converting all diffusion MRI scans  $\{d_i\}_{i=1}^n$ from DICOM to NIFTI format.
This is followed by diffusion tensor estimation and computation of Fractional Anisotropy (FA) maps. The JHU white-matter atlas is then registered to each subject’s FA image to obtain ROI parcellations, and mean FA values are extracted from $p$ predefined ROIs $[r_j]_{j=1}^p$(details provided in Section~\ref{sec:exp}). 
%
Subsequently we construct histogram-based representations of FA distributions within each ROI. Specifically, for each subject $s_i$ with diffusion scan $d_i$, an initially empty feature vector $\mathbf{H}_{s_i}$ is created. For every ROI $r_j$, we collect the set of FA values $\mathcal{V}_{i,j} \in [0,1]^{k_{ij}}$ within that region, where $k_{ij}$ are number of voxels in that region. The interval $[0,1]$ is partitioned into $q$ equal-width bins $\mathcal{B} = \{b_1, b_2, \ldots, b_q\}$ to balance granularity and robustness. A normalized histogram is then computed as
$
h_{i,j} = \frac{\mathrm{Histogram}(\mathcal{V}_{i,j}, \mathcal{B})}{k_{ij}},
$
where $h_{i,j}$ represents the empirical distribution of FA values within ROI $r_j$ for subject $s_i$. Each $h_{i,j}$ is appended to $H_{s_i}$, yielding a concatenated feature vector comprising histogram-based FA representations across all ROIs. Let $H = [H_{s_i}]_{i=1}^{n} \in \mathcal{R}^{n\times pq}$ and adjacency matrix $A \in [0,1]^{n\times n}$ with $A_{ij} = (H_{s_i}^TH_{s_j} > \alpha)$. Here $\alpha$ is edge selection threshold.
%
We observed although GNN based methods have out-performed on Diffusion MRI classification\cite{gao2025graph}, however they suffer from oversmoothing and oversquashing problems\cite{zhaopairnorm}, which is well addressed by the variants of the GNNs \cite{gasteiger2018predict} \cite{lin2021mesh}. Hence, our proposed method include ARMA Convolution network \cite{bianchi2021graph} combined with reconstruction loss as regularizer \cite{wang2017mgae}.  ARMA produces rational filters (ratio of polynomials), which are more expressive than the polynomial filters used in GCN as mentioned in equation \ref{eq:arma_filter}.
\begin{equation}
X = 
\left(
I + \sum_{k=1}^{K} q_k L^k
\right)^{-1}
\left(
\sum_{k=0}^{K-1} p_k L^k
\right) H.
\label{eq:arma_filter}
\end{equation}
Different orders $(K)$ of the numerator and denominator in~\eqref{eq:arma_filter} are trivially obtained by setting some coefficients to zero. 
By setting $q_k = 0$ for every $k$, one recovers a polynomial filter, which corresponds to the moving-average (MA) term of the model. 
The inclusion of the additional autoregressive (AR) term, encoded by these coefficients, makes the ARMA model robust to noise and allows it to capture longer dynamics on the graph, since $X$ depends, in turn, on several steps of propagation of the node features. 
This is the key to capturing longer dependencies and more global structures on the graph compared to a polynomial filter of the same degree without exploding the number of parameters.
$X$ is approximated as 
\begin{equation}
X_r^{(l+1)} = \sigma(\tilde{A}X_r^{l}W_{(r)}^{(l)} + HV_{(r)}^{(l)}),
\label{eq:Arma_Conv}
\end{equation}
where $W_{(r)}^{(l)}, V_{(r)}^{(l)}$ are the learnable parameters at $l^{th}$ layer and $r^{th}$ row stack, $H_{s_i}$ is the initial node features. Here $\tilde{A} = D^{\frac{-1}{2}}\hat{A}D^{\frac{-1}{2}}$ and $\hat{A} = A + I_n$, where $D \in \mathbb{R}^{n\times n}$ is a diagonal degree matrix with $D_{ii} = d_i$, in which $d_i$ represents degree of $i^{th}$ node, and $I_n$ is $n\times n$ identity matrix. 
%
%
Our proposed decoder architecture is based on decoding framework of \cite{kim2024gala} but instead of GCN based Laplacian sharpening associated with decoder, we used MLP layer.
As decoder is deployed to reconstruct back the original features, hence our proposed MLP layers will be implicitly learning the intended sharpening effect.
%
\section{Experiments and Results}
\label{sec:exp}
\noindent
\noindent
We evaluate the proposed ARMARecon framework on diffusion MRI data from the Alzheimer's Disease Neuroimaging Initiative (ADNI) and the Neuroimaging in Frontotemporal Dementia (NIFD) datasets, encompassing subjects with varying cognitive impairment levels. The analysis targets white matter regions known to exhibit microstructural degeneration in Alzheimer's disease (AD), quantified through Fractional Anisotropy (FA) derived from diffusion tensor imaging. The selected nine regions are well-established in prior literature for their sensitivity to early AD-related changes. Specifically, the \textit{corpus callosum}, a major interhemispheric pathway, shows FA reductions linked to axonal and myelin loss preceding atrophy and cognitive decline~\cite{walterfang2014shape}.
The \textit{fornix}, a key hippocampal efferent tract, demonstrates early FA decline predictive of MCI-to-AD conversion and correlating with episodic memory deficits~\cite{nowrangi2015fornix}.
The \textit{right corticospinal tract} and \textit{right cerebral peduncle} show FA decreases reflecting axonal disorganization and executive dysfunction even in the absence of white matter hyperintensities~\cite{teipel2014fractional}.
The \textit{right anterior limb of the internal capsule} exhibits FA reductions associated with processing speed and working memory deficits~\cite{madden2008cerebral, mielke2009regionally}. Finally, the \textit{cingulum bundle}, particularly its hippocampal segment, shows FA reduction strongly correlated with episodic memory decline, highlighting differential anterior–posterior vulnerability patterns in AD progression~\cite{zhuang2010white}.
The experiments focused on diffusion MRI datasets featuring subjects with different cognitive conditions. This included Alzheimer’s disease (AD), mild cognitive impairment (MCI), and cognitively normal (CN) controls. Selected cohorts of ADNI dataset includes 388 subjects: 133 CN, 167 MCI, and 88 AD. This data includes scans with 46, 54, and 55 diffusion directions. We also tested the model on a NIFD cohort, which included 98 FTD patients and 48 healthy controls scanned with 120 diffusion-weighted directions.

For all subjects, we computed fractional anisotropy (FA) maps using the DIPY library~\cite{garyfallidis2014dipy}. From each of the nine selected white matter ROIs, we extracted FA histograms using $q=20$ bins, giving a total of $p=9$ ROIs. This produced a 180-dimensional feature vector for each subject. These histogram features were then used to build subject-level graphs based on cosine similarity, with edge sparsity controlled by the threshold parameter $\alpha$.

The ARMARecon model used the ReLU activation function and a dropout rate of 0.25 to reduce overfitting. We optimized the model with the Adam optimizer, setting the learning rate to $1\times10^{-4}$ and weight decay to $1\times10^{-4}$. In the ARMA convolutional layer, we set the number of stacks and the number of layers to 1 ($\text{num\_stacks}=1$, $\text{num\_layers}=1$) to maintain a compact architecture and prevent over-parameterization while ensuring efficient feature propagation across the graph.
We trained it for 2000 epochs using a 20-fold stratified cross-validation scheme. The model showed sensitivity to $\alpha$, with optimal values of 0.92 for ADNI (CN vs. MCI), 0.8 for ADNI (CN vs AD), and 0.5 for NIFD. These values reflected the different graph density requirements for each dataset. The reconstruction loss coefficient ($\lambda_{\text{recon}}$) improved classification performance, through trade-off between supervised and self-reconstruction objectives.
To get better results, we adjusted the reconstruction weight $\lambda_{\text{recon}}$ for each dataset and train-test split separately. For the 90-10 split, the best values were 0.9 for ADNI (CN-MCI), 0.9 for ADNI (CN-AD), and 0.001 for NIFD. For the 70-30 split, the optimal values were 0.005 for CN-MCI, 2 for CN-AD, and 0.001 for NIFD. In the 50-50 split, the best settings were 0.001 for CN-MCI, 0.9 for CN-AD, and 0.001 for NIFD. These changes show how different datasets affect graph density and label imbalance. This highlights the need to adjust the reconstruction regularization for robust and precise classification.
We performed evaluation across multiple train–test splits (90–10, 70–30, and 50–50) to ensure robustness of all models at different levels of availability of training labels. 
Table \ref{table:split50_cv_updated_auc} showed that \textbf{ARMARecon} consistently outperformed all baseline models. These included Support Vector Classifier (SVC)~\cite{cortes1995support}, XGBoost~\cite{chen2016xgboost}, Random Forest~\cite{breiman2001random}, Multilayer Perceptron (MLP)~\cite{rumelhart1986learning}, Graph Convolutional Network (GCN)~\cite{kipf2017semi}, Graph Attention Network (GAT)~\cite{veličković2018graphattentionnetworks}, and Chebyshev Network (ChebNet)~\cite{defferrard2016convolutional}. It also surpassed the ARMA Graph Network baseline~\cite{bianchi2021graph}, demonstrating the effectiveness of integrating a reconstruction-driven regularization strategy. The proposed model achieved the highest accuracy, precision, recall, F1-score, and AUC across both the ADNI and NIFD cohorts. These results confirm that FA-derived graph representations effectively capture microstructural changes related to cognitive decline. All experiments were conducted on an NVIDIA RTX A4000 GPU, ensuring reproducible and efficient model optimization.

\begin{table*}[ht!]
\centering
\setlength{\tabcolsep}{6pt}
\renewcommand{\arraystretch}{1.15}

\resizebox{\textwidth}{!}{
\begin{tabular}{l|ccccc|ccccc|ccccc}
\hline
\textbf{(A) Methods} &
\multicolumn{5}{c|}{\textbf{ADNI (CN vs MCI)}} &
\multicolumn{5}{c|}{\textbf{ADNI (CN vs AD)}} &
\multicolumn{5}{c}{\textbf{NIFD}} \\
 & Acc & Prec & Rec & F1 & AUC & Acc & Prec & Rec & F1 & AUC & Acc & Prec & Rec & F1 & AUC \\ \hline
SVC~\cite{cortes1995support} & 0.807 ± 0.06 & 0.855 ± 0.07 & 0.803 ± 0.09 & 0.824 ± 0.06 & 0.877 ± 0.05 & 0.863 ± 0.06 & 0.838 ± 0.10 & 0.822 ± 0.11 & 0.823 ± 0.08 & 0.895 ± 0.06 & 0.720 ± 0.11 & 0.785 ± 0.09 & 0.810 ± 0.09 & 0.794 ± 0.08 & 0.789 ± 0.10 \\
XGBoost~\cite{chen2016xgboost} & 0.813 ± 0.07 & 0.849 ± 0.09 & 0.829 ± 0.08 & 0.835 ± 0.06 & 0.906 ± 0.05 & 0.824 ± 0.07 & 0.792 ± 0.11 & 0.761 ± 0.12 & 0.770 ± 0.10 & 0.851 ± 0.09 & 0.727 ± 0.11 & 0.779 ± 0.08 & 0.830 ± 0.12 & 0.800 ± 0.09 & 0.807 ± 0.10 \\
MLPClassifier~\cite{rumelhart1986learning} & 0.813 ± 0.07 & 0.836 ± 0.08 & 0.844 ± 0.08 & 0.837 ± 0.06 & 0.889 ± 0.06 & 0.859 ± 0.07 & 0.829 ± 0.11 & 0.822 ± 0.12 & 0.819 ± 0.09 & 0.898 ± 0.07 & 0.717 ± 0.10 & 0.819 ± 0.09 & 0.803 ± 0.31 & 0.776 ± 0.09 & 0.803 ± 0.09 \\
Random Forest~\cite{breiman2001random} & 0.815 ± 0.05 & 0.858 ± 0.08 & 0.821 ± 0.10 & 0.833 ± 0.05 & 0.913 ± 0.04 & 0.820 ± 0.09 & 0.798 ± 0.14 & 0.744 ± 0.15 & 0.761 ± 0.12 & 0.870 ± 0.08 & 0.770 ± 0.08 & 0.778 ± 0.06 & 0.925 ± 0.07 & 0.843 ± 0.05 & 0.820 ± 0.10 \\
AE-GCN ~\cite{ma2021aegcn} & 0.808 ± 0.06 & 0.848 ± 0.06 & 0.815 ± 0.10 & 0.826 ± 0.06 & 0.886 ± 0.05 & 0.808 ± 0.06 & 0.826 ± 0.06 & 0.815 ± 0.10 & 0.826 ± 0.06 & 0.885 ± 0.05 & 0.730 ± 0.08 & 0.692 ± 0.36 & 0.290 ± 0.17 & 0.394 ± 0.22 & 0.808 ± 0.09 \\
GCN~\cite{kipf2017semi} & 0.747 ± 0.09 & 0.780 ± 0.09 & 0.777 ± 0.12 & 0.774 ± 0.09 & 0.812 ± 0.09 & 0.713 ± 0.09 & 0.722 ± 0.20 & 0.589 ± 0.25 & 0.593 ± 0.16 & 0.811 ± 0.10 & 0.680 ± 0.07 & 0.782 ± 0.08 & 0.745 ± 0.13 & 0.752 ± 0.07 & 0.774 ± 0.08 \\
GCN (Recon) & 0.975 ± 0.03 & 0.975 ± 0.03 & 0.982 ± 0.03 & 0.978 ± 0.02 & 0.993 ± 0.01 & 0.863 ± 0.09 & 0.879 ± 0.13 & 0.783 ± 0.18 & 0.812 ± 0.13 & 0.926 ± 0.07 & 0.840 ± 0.10 & 0.918 ± 0.07 & 0.846 ± 0.16 & 0.869 ± 0.09 & 0.929 ± 0.07 \\
GAT~\cite{veličković2018graphattentionnetworks} & 0.777 ± 0.02 & 0.787 ± 0.03 & 0.826 ± 0.05 & 0.804 ± 0.02 & 0.849 ± 0.02 & 0.757 ± 0.07 & 0.700 ± 0.12 & 0.728 ± 0.13 & 0.699 ± 0.08 & 0.839 ± 0.08 & 0.697 ± 0.09 & 0.793 ± 0.09 & 0.750 ± 0.12 & 0.765 ± 0.07 & 0.759 ± 0.10 \\
GAT (Recon) & 0.955 ± 0.07 & 0.976 ± 0.04 & 0.944 ± 0.10 & 0.957 ± 0.07 & 0.984 ± 0.02 & 0.983 ± 0.03 & 0.976 ± 0.05 & 0.983 ± 0.04 & 0.978 ± 0.03 & 0.993 ± 0.02 & \textbf{0.983 ± 0.04} & \textbf{0.990 ± 0.04} & \textbf{0.960 ± 0.08} & \textbf{0.973 ± 0.06} & \textbf{0.992 ± 0.03} \\
ChebNet~\cite{defferrard2016convolutional} & 0.733 ± 0.09 & 0.776 ± 0.09 & 0.753 ± 0.13 & 0.759 ± 0.09 & 0.804 ± 0.08 & 0.783 ± 0.06 & 0.716 ± 0.09 & 0.761 ± 0.09 & 0.733 ± 0.07 & 0.849 ± 0.06 & 0.537 ± 0.10 & 0.747 ± 0.11 & 0.460 ± 0.15 & 0.558 ± 0.13 & 0.689 ± 0.12 \\
ChebNet (Recon) & 0.978 ± 0.02 & 0.984 ± 0.03 & 0.979 ± 0.03 & 0.981 ± 0.02 & 0.995 ± 0.01 & 0.994 ± 0.02 & 0.990 ± 0.03 & 0.995 ± 0.02 & 0.992 ± 0.02 & 0.996 ± 0.02 & 0.980 ± 0.03 & 0.982 ± 0.04 & 0.990 ± 0.03 & 0.985 ± 0.02 & 1.000 ± 0.00 \\
ARMA~\cite{bianchi2021graph} & 0.793 ± 0.08 & 0.827 ± 0.08 & 0.812 ± 0.12 & 0.814 ± 0.08 & 0.861 ± 0.09 & 0.822 ± 0.08 & 0.783 ± 0.13 & 0.789 ± 0.09 & 0.779 ± 0.08 & 0.884 ± 0.07 & 0.623 ± 0.10 & 0.815 ± 0.11 & 0.570 ± 0.14 & 0.660 ± 0.11 & 0.743 ± 0.11 \\
\textbf{ARMA (Recon)} & \textbf{0.983 ± 0.02} & \textbf{0.989 ± 0.03} & \textbf{0.982 ± 0.03} & \textbf{0.985 ± 0.02} & \textbf{0.997 ± 0.01} & \textbf{0.997 ± 0.01} & \textbf{1.000 ± 0.00} & \textbf{0.993 ± 0.03} & \textbf{0.997 ± 0.01} & \textbf{0.999 ± 0.00} & 0.977 ± 0.04 & 0.978 ± 0.05 & 0.990 ± 0.03 & 0.983 ± 0.03 & 1.000 ± 0.00 \\
\hline
\end{tabular}}
\label{table:split90_cv_corrected_bold}
\end{table*}

\begin{table*}[ht!]
\centering
\setlength{\tabcolsep}{6pt}
\renewcommand{\arraystretch}{1.15}
\resizebox{\textwidth}{!}{
\begin{tabular}{l|ccccc|ccccc|ccccc}
\hline
\textbf{(B) Methods} &
\multicolumn{5}{c|}{\textbf{ADNI (CN vs MCI)}} &
\multicolumn{5}{c|}{\textbf{ADNI (CN vs AD)}} &
\multicolumn{5}{c}{\textbf{NIFD}} \\
 & Acc & Prec & Rec & F1 & AUC & Acc & Prec & Rec & F1 & AUC & Acc & Prec & Rec & F1 & AUC \\ \hline
SVC~\cite{cortes1995support} & 0.801 ± 0.02 & 0.824 ± 0.03 & 0.818 ± 0.04 & 0.820 ± 0.02 & 0.864 ± 0.02 & 0.795 ± 0.02 & 0.782 ± 0.02 & 0.678 ± 0.06 & 0.725 ± 0.04 & 0.854 ± 0.02 & 0.727 ± 0.04 & 0.783 ± 0.03 & 0.833 ± 0.06 & 0.806 ± 0.03 & 0.803 ± 0.05 \\
XGBoost~\cite{chen2016xgboost} & 0.803 ± 0.03 & 0.823 ± 0.03 & 0.825 ± 0.05 & 0.822 ± 0.03 & 0.882 ± 0.02 & 0.810 ± 0.04 & 0.813 ± 0.08 & 0.700 ± 0.08 & 0.748 ± 0.06 & 0.848 ± 0.03 & 0.734 ± 0.06 & 0.781 ± 0.04 & 0.850 ± 0.07 & 0.813 ± 0.04 & 0.790 ± 0.06 \\
MLPClassifier~\cite{rumelhart1986learning} & 0.803 ± 0.03 & 0.816 ± 0.04 & 0.838 ± 0.03 & 0.826 ± 0.02 & 0.868 ± 0.02 & 0.796 ± 0.03 & 0.748 ± 0.04 & 0.742 ± 0.05 & 0.744 ± 0.04 & 0.857 ± 0.02 & 0.747 ± 0.05 & 0.835 ± 0.04 & 0.787 ± 0.07 & 0.808 ± 0.04 & 0.817 ± 0.05 \\
Random Forest~\cite{breiman2001random} & 0.815 ± 0.03 & 0.834 ± 0.03 & 0.835 ± 0.04 & 0.834 ± 0.03 & 0.893 ± 0.02 & 0.786 ± 0.02 & 0.790 ± 0.03 & 0.637 ± 0.06 & 0.704 ± 0.04 & 0.854 ± 0.02 & 0.744 ± 0.04 & 0.770 ± 0.03 & 0.895 ± 0.06 & 0.826 ± 0.03 & 0.809 ± 0.06 \\
AE-GCN ~\cite{ma2021aegcn} & 0.793 ± 0.03 & 0.835 ± 0.04 & 0.787 ± 0.05 & 0.809 ± 0.03 & 0.866 ± 0.02 & 0.732 ± 0.04 & 0.801 ± 0.08 & 0.448 ± 0.08 & 0.572 ± 0.07 & 0.831 ± 0.04 & 0.724 ± 0.05 & 0.705 ± 0.26 & 0.243 ± 0.13 & 0.344 ± 0.15 & 0.756 ± 0.07 \\
GCN~\cite{kipf2017semi} & 0.751 ± 0.03 & 0.776 ± 0.04 & 0.781 ± 0.06 & 0.776 ± 0.03 & 0.818 ± 0.04 & 0.713 ± 0.08 & 0.735 ± 0.16 & 0.617 ± 0.20 & 0.627 ± 0.08 & 0.821 ± 0.03 & 0.699 ± 0.07 & 0.804 ± 0.07 & 0.752 ± 0.12 & 0.769 ± 0.06 & 0.775 ± 0.07 \\
GCN (Recon) & 0.932 ± 0.04 & 0.953 ± 0.04 & 0.927 ± 0.05 & 0.939 ± 0.03 & 0.965 ± 0.04 & 0.831 ± 0.05 & 0.859 ± 0.10 & 0.720 ± 0.13 & 0.770 ± 0.08 & 0.895 ± 0.03 & 0.828 ± 0.05 & 0.771 ± 0.11 & 0.689 ± 0.17 & 0.710 ± 0.11 & 0.900 ± 0.04 \\
GAT~\cite{veličković2018graphattentionnetworks} & 0.773 ± 0.04 & 0.792 ± 0.04 & 0.804 ± 0.06 & 0.797 ± 0.04 & 0.848 ± 0.03 & 0.769 ± 0.04 & 0.703 ± 0.06 & 0.756 ± 0.08 & 0.725 ± 0.04 & 0.841 ± 0.03 & 0.708 ± 0.04 & 0.796 ± 0.05 & 0.778 ± 0.09 & 0.783 ± 0.04 & 0.770 ± 0.05 \\
GAT (Recon) & 0.920 ± 0.03 & 0.931 ± 0.03 & 0.926 ± 0.06 & 0.927 ± 0.03 & 0.959 ± 0.02 & 0.935 ± 0.03 & 0.917 ± 0.04 & 0.924 ± 0.05 & 0.920 ± 0.03 & 0.962 ± 0.02 & 0.906 ± 0.04 & 0.864 ± 0.08 & 0.843 ± 0.07 & 0.851 ± 0.06 & 0.946 ± 0.03 \\
ChebNet~\cite{defferrard2016convolutional} & 0.757 ± 0.04 & 0.784 ± 0.04 & 0.780 ± 0.05 & 0.781 ± 0.04 & 0.825 ± 0.03 & 0.756 ± 0.04 & 0.677 ± 0.04 & 0.759 ± 0.07 & 0.714 ± 0.05 & 0.832 ± 0.04 & 0.605 ± 0.06 & 0.824 ± 0.07 & 0.537 ± 0.06 & 0.648 ± 0.06 & 0.723 ± 0.07 \\
ChebNet (Recon) & 0.931 ± 0.02 & 0.937 ± 0.03 & 0.940 ± 0.04 & 0.938 ± 0.02 & 0.978 ± 0.01 & 0.943 ± 0.02 & 0.920 ± 0.04 & 0.944 ± 0.04 & 0.931 ± 0.03 & 0.983 ± 0.01 & 0.913 ± 0.04 & 0.862 ± 0.07 & 0.868 ± 0.07 & 0.863 ± 0.06 & 0.973 ± 0.02 \\
ARMA~\cite{bianchi2021graph} & 0.803 ± 0.03 & 0.824 ± 0.04 & 0.824 ± 0.05 & 0.823 ± 0.03 & 0.875 ± 0.03 & 0.802 ± 0.05 & 0.747 ± 0.07 & 0.774 ± 0.09 & 0.758 ± 0.06 & 0.870 ± 0.04 & 0.663 ± 0.07 & 0.830 ± 0.07 & 0.638 ± 0.09 & 0.718 ± 0.07 & 0.771 ± 0.06 \\
\textbf{ARMA (Recon)} & \textbf{0.939 ± 0.02} & \textbf{0.946 ± 0.02} & \textbf{0.945 ± 0.03} & \textbf{0.945 ± 0.02} & \textbf{0.983 ± 0.01} & \textbf{0.952 ± 0.02} & \textbf{0.935 ± 0.04} & \textbf{0.950 ± 0.04} & \textbf{0.942 ± 0.03} & \textbf{0.987 ± 0.01} & \textbf{0.918 ± 0.03} & \textbf{0.891 ± 0.06} & \textbf{0.850 ± 0.08} & \textbf{0.868 ± 0.05} & \textbf{0.979 ± 0.01} \\
\hline
\end{tabular}}
\label{table:split70_cv_corrected_bold}
\end{table*}

\begin{table*}[ht!]
\centering
\setlength{\tabcolsep}{6pt}
\renewcommand{\arraystretch}{1.15}
\resizebox{\textwidth}{!}{
\begin{tabular}{l|ccccc|ccccc|ccccc}
\hline
\textbf{(C) Methods} &
\multicolumn{5}{c|}{\textbf{ADNI (CN vs MCI)}} &
\multicolumn{5}{c|}{\textbf{ADNI (CN vs AD)}} &
\multicolumn{5}{c}{\textbf{NIFD}} \\
 & Acc & Prec & Rec & F1 & AUC & Acc & Prec & Rec & F1 & AUC & Acc & Prec & Rec & F1 & AUC \\ \hline

SVC~\cite{cortes1995support}
& 0.796 ± 0.02 & 0.819 ± 0.03 & 0.816 ± 0.05 & 0.816 ± 0.02 & 0.862 ± 0.02
& 0.796 ± 0.03 & 0.779 ± 0.05 & 0.690 ± 0.07 & 0.729 ± 0.04 & 0.863 ± 0.02
& 0.717 ± 0.03 & 0.760 ± 0.03 & 0.851 ± 0.05 & 0.801 ± 0.02 & 0.767 ± 0.04 \\

XGBoost~\cite{chen2016xgboost}
& 0.807 ± 0.03 & 0.820 ± 0.04 & 0.841 ± 0.04 & 0.829 ± 0.02 & 0.874 ± 0.02
& 0.784 ± 0.03 & 0.768 ± 0.05 & 0.664 ± 0.07 & 0.710 ± 0.05 & 0.840 ± 0.03
& 0.713 ± 0.04 & 0.759 ± 0.03 & 0.841 ± 0.04 & 0.797 ± 0.03 & 0.743 ± 0.06 \\

MLPClassifier~\cite{rumelhart1986learning}
& 0.795 ± 0.02 & 0.808 ± 0.03 & 0.831 ± 0.04 & 0.818 ± 0.02 & 0.862 ± 0.02
& 0.802 ± 0.03 & 0.767 ± 0.04 & 0.732 ± 0.07 & 0.747 ± 0.05 & 0.861 ± 0.02
& 0.721 ± 0.03 & 0.806 ± 0.05 & 0.775 ± 0.04 & 0.788 ± 0.02 & 0.779 ± 0.04 \\

Random Forest~\cite{breiman2001random}
& 0.804 ± 0.02 & 0.823 ± 0.04 & 0.830 ± 0.05 & 0.825 ± 0.02 & 0.885 ± 0.02
& 0.782 ± 0.03 & 0.788 ± 0.05 & 0.629 ± 0.07 & 0.697 ± 0.05 & 0.863 ± 0.03
& 0.724 ± 0.04 & 0.739 ± 0.03 & 0.912 ± 0.05 & 0.816 ± 0.03 & 0.793 ± 0.06 \\

AE-GCN~\cite{ma2021aegcn}
& 0.793 ± 0.07 & 0.839 ± 0.09 & 0.803 ± 0.10 & 0.814 ± 0.07 & 0.870 ± 0.05
& 0.726 ± 0.03 & 0.785 ± 0.05 & 0.439 ± 0.06 & 0.561 ± 0.06 & 0.826 ± 0.03
& 0.730 ± 0.07 & 0.580 ± 0.42 & 0.280 ± 0.21 & 0.365 ± 0.27 & 0.791 ± 0.08 \\

GCN~\cite{kipf2017semi}
& 0.754 ± 0.03 & 0.769 ± 0.04 & 0.801 ± 0.05 & 0.783 ± 0.03 & 0.822 ± 0.03
& 0.722 ± 0.06 & 0.705 ± 0.13 & 0.654 ± 0.18 & 0.644 ± 0.10 & 0.796 ± 0.03
& 0.662 ± 0.05 & 0.768 ± 0.05 & 0.720 ± 0.09 & 0.738 ± 0.05 & 0.714 ± 0.04 \\

GCN (Recon)
& 0.887 ± 0.03 & 0.903 ± 0.04 & 0.895 ± 0.04 & 0.898 ± 0.03 & 0.935 ± 0.02
& 0.804 ± 0.07 & 0.802 ± 0.12 & 0.741 ± 0.11 & 0.756 ± 0.05 & 0.872 ± 0.04
& 0.770 ± 0.04 & 0.838 ± 0.06 & 0.827 ± 0.10 & 0.826 ± 0.04 & 0.825 ± 0.04 \\

GAT~\cite{veličković2018graphattentionnetworks}
& 0.730 ± 0.04 & 0.753 ± 0.06 & 0.787 ± 0.12 & 0.760 ± 0.05 & 0.807 ± 0.04
& 0.775 ± 0.04 & 0.714 ± 0.06 & 0.744 ± 0.07 & 0.726 ± 0.05 & 0.849 ± 0.03
& 0.686 ± 0.05 & 0.773 ± 0.05 & 0.760 ± 0.08 & 0.764 ± 0.04 & 0.738 ± 0.06 \\

GAT (Recon)
& 0.886 ± 0.02 & 0.905 ± 0.03 & 0.890 ± 0.04 & 0.896 ± 0.02 & 0.934 ± 0.02
& 0.896 ± 0.03 & 0.861 ± 0.04 & 0.886 ± 0.05 & 0.872 ± 0.04 & 0.941 ± 0.02
& 0.851 ± 0.05 & 0.783 ± 0.08 & 0.769 ± 0.08 & 0.773 ± 0.07 & 0.895 ± 0.04 \\

ChebNet~\cite{defferrard2016convolutional}
& 0.748 ± 0.04 & 0.785 ± 0.03 & 0.753 ± 0.05 & 0.768 ± 0.04 & 0.826 ± 0.03
& 0.763 ± 0.04 & 0.682 ± 0.05 & 0.770 ± 0.06 & 0.723 ± 0.05 & 0.842 ± 0.04
& 0.624 ± 0.04 & 0.830 ± 0.06 & 0.557 ± 0.05 & 0.665 ± 0.04 & 0.729 ± 0.06 \\

ChebNet (Recon)
& 0.899 ± 0.01 & 0.912 ± 0.02 & 0.907 ± 0.03 & 0.909 ± 0.01 & 0.954 ± 0.01
& 0.903 ± 0.03 & 0.869 ± 0.04 & 0.893 ± 0.05 & 0.880 ± 0.04 & 0.958 ± 0.02
& 0.853 ± 0.02 & 0.905 ± 0.03 & 0.877 ± 0.05 & 0.889 ± 0.02 & 0.934 ± 0.02 \\

ARMA~\cite{bianchi2021graph}
& 0.800 ± 0.03 & 0.819 ± 0.02 & 0.822 ± 0.04 & 0.820 ± 0.03 & 0.868 ± 0.02
& 0.793 ± 0.04 & 0.731 ± 0.05 & 0.771 ± 0.06 & 0.749 ± 0.04 & 0.869 ± 0.03
& 0.649 ± 0.05 & 0.818 ± 0.06 & 0.618 ± 0.06 & 0.702 ± 0.04 & 0.743 ± 0.04 \\

\textbf{ARMA (Recon)}
& \textbf{0.904 ± 0.02} & \textbf{0.911 ± 0.02} & \textbf{0.919 ± 0.03} & \textbf{0.914 ± 0.02} & \textbf{0.959 ± 0.01}
& \textbf{0.909 ± 0.02} & \textbf{0.888 ± 0.03} & \textbf{0.887 ± 0.03} & \textbf{0.887 ± 0.02} & \textbf{0.967 ± 0.01}
& \textbf{0.860 ± 0.03} & \textbf{0.887 ± 0.04} & \textbf{0.910 ± 0.04} & \textbf{0.897 ± 0.02} & \textbf{0.939 ± 0.02} \\

\hline
\end{tabular}}
\caption{Performance comparison (mean ± std) for (A)90-10 (B) 70–30 (C) 50-50 train–test split with 20-fold cross-validation across datasets. Highest values per dataset are in \textbf{bold}.}
\label{table:split50_cv_updated_auc}
\end{table*}



\section{Conclusions}
\noindent
Direct classification of raw diffusion MRI data remains computationally intensive and impractical for studies with limited samples such as neurodegenerative diseases. Our approach overcomes this limitation through compact, biologically informed histogram-based features that capture key microstructural alterations while reducing dimensionality. This framework enables scalable, interpretable, and privacy-preserving analysis across multi-institutional settings.
These histogram features are further fed to the ARMARecon network.The ARMA part overcomes the oversquashing issue faced by the GNN while reconstruction regularizer incorporated Laplacian sharpening which addresses oversmoothing problem.  

\section*{Acknowledgment}
Data used in the preparation of this article were obtained from the Alzheimer’s Disease Neuroimaging Initiative (ADNI) database (\url{adni.loni.usc.edu}) and the Neuroimaging in Frontotemporal Dementia (NIFD) study, also known as the Frontotemporal Lobar Degeneration Neuroimaging Initiative (FTLDNI) (\url{ida.loni.usc.edu}). 
The investigators within ADNI and NIFD contributed to the design and implementation of their respective studies and/or provided data, but did not participate in the analysis or writing of this report. 
A complete listing of ADNI investigators can be found at: ADNI Acknowledgement List (\url{http://adni.loni.usc.edu/wp-content/uploads/how_to_apply/ADNI_Acknowledgement_List.pdf}). 
Information about the NIFD (FTLDNI) study and its investigators is available at: NIFD Investigator List(\url{https://ida.loni.usc.edu/login.jsp?project=NIFD}).

\bibliographystyle{IEEEtran}
\bibliography{reference}
\end{document}